\documentclass[pdflatex,sn-mathphys-num]{sn-jnl}
\usepackage[T1]{fontenc}
\usepackage{graphicx}
\usepackage{multirow}
\usepackage{amsmath,amssymb,amsfonts}
\usepackage{amsthm}
\usepackage{mathrsfs}
\usepackage[title]{appendix}
\usepackage{xcolor}
\usepackage{textcomp}
\usepackage{manyfoot}
\usepackage{booktabs}
\usepackage{algorithm}
\usepackage{algorithmicx}
\usepackage{algpseudocode}
\usepackage{listings}
\usepackage{float}
\usepackage{makecell}

\theoremstyle{thmstyleone}

\theoremstyle{thmstyletwo}

\theoremstyle{thmstylethree}

\begin{document}

\title[RKO for Probability-Driven UAV Search]{Maximizing Detection Probability in UAV Search Missions using Random-Key Optimizers}

\author*[1]{\fnm{Antonio} \sur{Dantas}}\email{antonio.josivaldo@unifesp.br}

\author[2]{\fnm{Elcio} \sur{Shiguemori}}\email{elciohs@gmail.com }

\author[1]{\fnm{Antonio} \sur{Chaves}}\email{antonio.chaves@unifesp.br}

\affil*[1]{\orgdiv{Institute of Science and Technology}, \orgname{Federal University of Sao Paulo}, \orgaddress{\city{Sao Jose dos Campos}, \state{SP}, \country{Brazil}}}

\affil[2]{\orgdiv{Department of Aerospace Science and Technology}, \orgname{Institute for Advanced Studies}, \orgaddress{\city{Sao Jose dos Campos}, \state{SP}, \country{Brazil}}}

\abstract{This paper focuses on the stochastic and real-time constraints of Unmanned Aerial Vehicle (UAV) path planning in Search and Rescue (SAR) missions. We introduce a configurable data-driven and intelligent routing architecture that combines a closed-loop probability update with the Random-Key Optimizers (RKO). Our problem is mathematically modeled as a dynamic probabilistic variant of the Open Vehicle Routing Problem (OVRP), enabling the solver to switch adaptively between two objectives, thereby balancing probability maximization and resource minimization. The framework degrades elegantly under noisy priors but not under biased ones: a systematically biased map performs about half as well as flying with no prior at all. Computational tests show that the proposed scheme is scalable and competitive with single-path scenarios in terms of solution performance and serves as a strong decision guide for intelligent planning in SAR operations in unstable environments.}

\keywords{Search and Rescue, UAV, Random-Key Optimizers, Vehicle Routing Problem, Probability-Driven Routing.}

\maketitle

\section{Introduction}
\label{sec:intro}

Climate volatility is making Search and Rescue (SAR) missions more frequent and harder to plan. In floods, landslides and similar events the scenario changes during the search itself, so the routing problem is stochastic by nature and current search practices struggle with it. There have been recent large-scale occurrences of this type, such as floods in Asia that resulted in over 1,700 deaths, as well as the dam collapse and floods in Brazil. Events on this scale put a premium on how the search itself is run: the techniques and strategies behind it exist to keep people alive while the danger is still immediate \cite{i1}. 

Unmanned Aerial Vehicles (UAVs) have earned their place in this research because they reach difficult ground at low cost \cite{i2}; what holds them back is energy, and the route planning that has to live within it \cite{i3}. In hydrological disasters time is the resource that matters most \cite{i4}, so our interest is in automating the search: missions that run faster, and victims that wait less, whatever the conditions on the ground.

What the planner produces is an ordered sequence of waypoints for each UAV. Guidance, stabilization and waypoint following stay with the onboard autopilot, so the method does not depend on the airframe: any platform that flies waypoints can execute it.

The flight path planning problem for UAVs in SAR operations involves searching for one or more static targets with unknown positions, the probable locations of which are provided by a heat map (probability density). The mathematical modeling of this scenario depends on the relationship between mission objectives and operational limitations, differing from the classic Vehicle Routing Problem (VRP) \cite{i5}. In one formulation, when the aim is to minimize travel distance subject to the requirement of collecting a minimum amount of probabilities. Conversely, when the objective is to maximize the total probabilities collected subject to a maximum cost or time limit (such as battery autonomy). In both approaches, the probability of a region containing a target is treated as the node value, allowing the system to focus on the most favorable areas.

We adopt the Random-Key Optimizers (RKO) \cite{i6} framework\footnote{\url{https://github.com/RKO-solver}}, which uses a decoder to map random-key vectors into feasible solutions. Since the decoder is the only problem-specific piece, the same solver can run several metaheuristics side by side, like Simulated Annealing (SA), Iterated Local Search (ILS), Variable Neighborhood Search (VNS), Large Neighborhood Search (LNS), Biased Random-Key Genetic Algorithms (BRKGA), Particle Swarm Optimization (PSO) and the Greedy Randomized Adaptive Search Procedure (GRASP), all of them reading from, and writing to, a single elite pool.

In this context, this study proposes an intelligent system that repeatedly optimizes UAV routes on a discretized probabilistic search map. This results in a close-loop optimization where the routing plan and mission intelligence evolve simultaneously. The core assumption of this method is that it shortens the time to localize target. Autonomous flight control and onboard perception are outside the scope of this work, we address the mission-level decision of which cells to observe, in which order, and with which vehicle. Specifically, the contributions are summarized as follows:

\begin{itemize}
    \item \textbf{Two OVRP formulations with target revisits:} We introduce an Integer Linear Programming (ILP) formulation for UAV search and rescue as a multi-depot Open Vehicle Routing Problem (OVRP), defining two complementary models: maximizing cumulative detection probability under mission time limits ($Z_1$) and minimizing fleet travel distance under coverage guarantees ($Z_2$), both accounting for battery return reserves and bounded node revisitation ($[V_{\min}, V_{\max}]$).
    \item \textbf{Tailored RKO decoding architecture:} We develop a two-stage delimiter-based chromosome decoder that maps continuous random keys into feasible multi-UAV routes and revisit intensities.
    \item \textbf{Controlled assessment of in-flight replanning:} We conduct an ablation study separating the effects of solver budget and replanning cadence. 
    \item \textbf{Resilience under prior map imperfections:} We systematically evaluate the system under noisy and biased probability maps, demonstrating graceful degradation under random noise, but catastrophic failure under systematic bias, providing clear operational guidelines for search and rescue managers.
\end{itemize}

This paper is organized: Section 2 provides related work; Section 3 explains the problem description and mathematical formulation; Section 4 discusses the RKO-based solution approach; Section 5 details computational experiments; In final section, conclusions are shown.

\section{Related Work}
\label{sec:related}

UAVs have also been increasingly utilized in SAR operations due to their capacity to rapidly analyze large volumes of imagery, thereby assisting in the identification of patterns that may indicate the presence of individuals \cite{i2}. Research into SAR operations can increase  efficiency and aid in the rapid location of lost or missing persons in emergency situations, even in hazardous areas, reducing the risk of injury to rescuers and decreasing operational costs \cite{r2}.

The work in \cite{r3} initiated an investigation into continuous monitoring using the VNS metaheuristic. That study ran into two walls: its solutions stayed up to 26.9\% away from the optimum even in small scenarios, and the exact methods would not finish in acceptable time on the large ones. The authors left two directions open, adapting the approach to dynamic environments and allowing base stations with no UAV assigned to them. The work in \cite{r4} took up this line and added service priorities and limited energy through the IDIH-Reserve heuristic.

The Drone Swarm Routing Problem (DSRP) approaches the problem in a fully decentralized and on-line manner, also on the basis of multi-agent systems, in which UAVs collaborate through "V" formation flights to save energy and increase the number of victims detected, proving the effectiveness of the method in the Beirut explosion case \cite{r5}. This study represents an evolution of the fundamental model subsequently published in \cite{r6}, where the authors introduced the Probabilistic Drone Routing Problem (PDRP), focused on maximizing identification expectation via the Adaptive Large Neighborhood Search (ALNS) metaheuristic. Although it laid theoretical foundations, the work \cite{r6} presents some operational limitations by relying on a static heatmap that ignores real-time terrain updates and by lacking exact methods to assess the absolute quality of heuristic solutions. Furthermore, the approach requires an expected number of targets at each node and applies a deterministic offline planning algorithm to solve a problem within an inherently stochastic and probabilistic environment, generating fixed routes that lack the dynamic adaptability essential in disasters a gap that the proposal in \cite{r5} sought to bridge.

In \cite{r7}, the authors demonstrated that GRASP outperforms greedy and local search heuristics in terms of quality and consistency within short time windows, the work possesses operational limitations by assuming static, two-dimensional maps, ignoring both topography and obstacles as well as collision dynamics between aircraft. Additionally, the model oversimplifies reality by presupposing infallible target detection in visited areas and disregarding the kinematic maneuverability constraints of the UAVs, in addition to relying on fixed parameter calibration that may not adapt to the variability of different disaster scenarios.

Existing literature on UAV routing often focuses on complete coverage strategies, prioritizing the exhaustive scanning of the entire area of interest. Motivated by the successes in solving highly complex problems and the ability of the RKO framework to handle different decoders, this work also focuses on the application of RKO using the decoding strategy proposed by \cite{r8}. Further, the majority of these studies rely on static maps, operating under the assumption that the environment and the target probability distribution remain invariant throughout the mission execution.

Recent robotics work addresses UAV search through multi-agent reinforcement learning \cite{r9} and evolutionary cooperative area search \cite{r10}, while in-flight re-planning has been demonstrated and field-validated for multi-UAV coverage \cite{r11}. These contributions establish that adaptive search is feasible, but may not isolate the adaptation itself, since the iterative system is not compared against the same configuration without feedback under a controlled budget. This work supplies both missing pieces: an exact formulation paired with the heuristic, so solution quality is reported as a gap rather than a ranking, and a controlled measurement of the value of iteration.

\section{Problem Description and Mathematical Formulation}
\label{sec:model}

The central problem addressed in this work consists of the efficient coordination of a UAV fleet to locate victims in a disaster-affected geographic region within a critical time horizon. Unlike conventional cargo transportation, where destinations are known deterministically, in SAR operations the final destination is latent. This fundamental uncertainty transforms the optimization task into a Probability-Driven Routing challenge, merging the complexities of probabilistic search with region coverage requirements.

Associated with the region is a probability of target presence, derived from a probability map (or heat map). This map is constructed from preliminary data, such as the Last Known Position (LKP), terrain topography, satellite imagery, and historical data of similar occurrences. The difficulty lies in the fact that these probabilities are not uniform; some areas are much more promising than others but may be geographically distant or difficult to access \cite{p1}.

\begin{figure}[htbp]
\centerline{
\includegraphics[width=\textwidth]{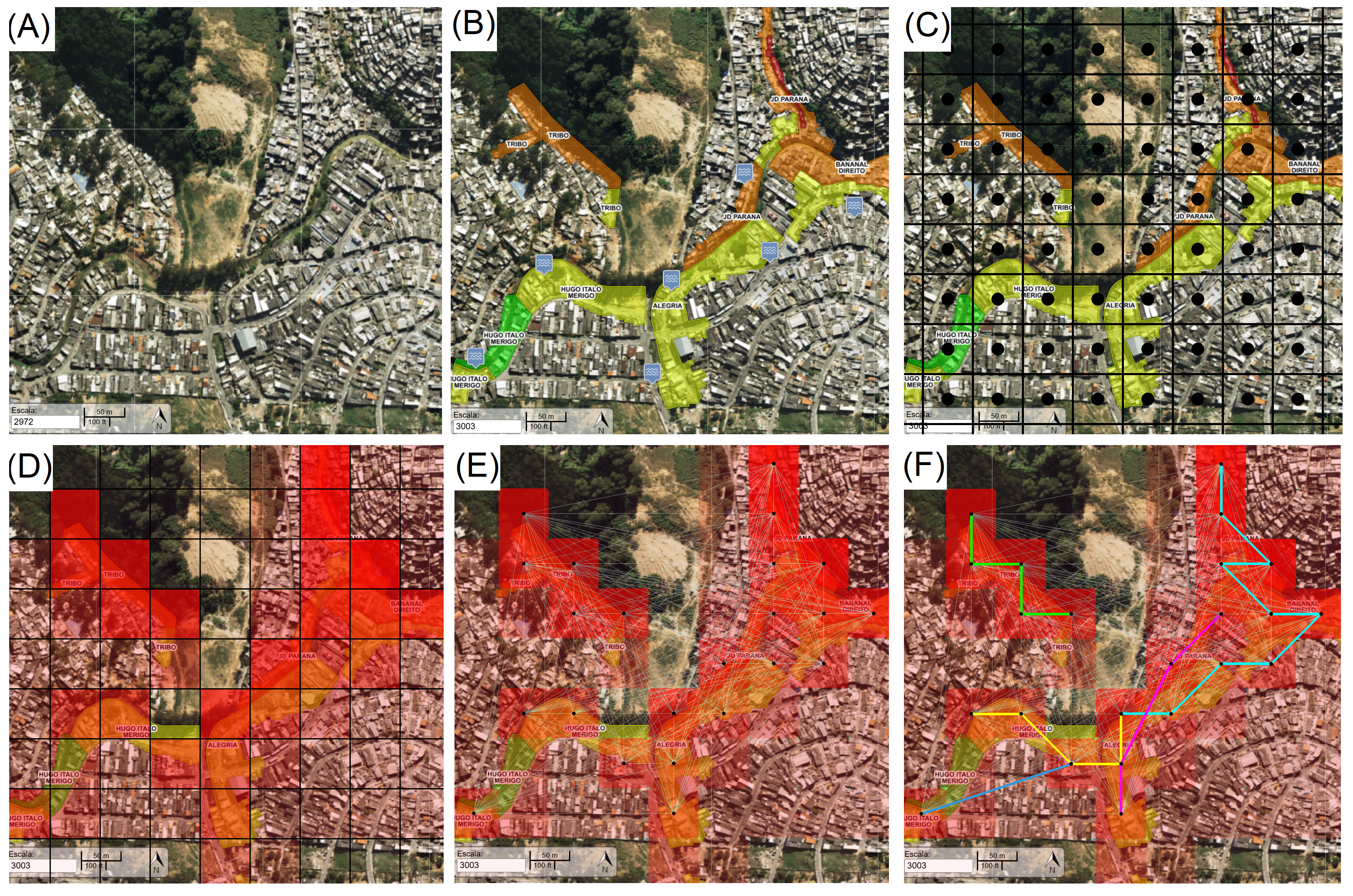}}
\caption{Construction of a search instance for a region of São Paulo. (A) Satellite view of the area. (B) Hydrological risk zones from GeoSampa, with the flood-risk markers. (C) Discretization into grid cells; each dot is a candidate point of interest. (D) Risk weights assigned to the cells, darker red meaning higher probability of target presence. (E) Feasible flight corridors connecting the points of interest. (F) Routes obtained after optimization, one color per UAV.} \label{mapa} \end{figure}

The region of interest must possess information regarding the probability of victim identification in specific areas. In the example shown in Fig.~\ref{mapa}, the area (A) and hydrological risk information was obtained from GeoSampa \footnote{The images and data are obtained from Digital Map of the City of São Paulo, available at https://geosampa.prefeitura.sp.gov.br.} (B), and the region is mapped and divided into a discrete grid of cells (C). Risk probabilities are weighted and associated with these cells (D). Each cell represents an area that can be scanned by a UAV in a single pass or observation, if selected, it is then determined to be a Point of Interest (POI). The dimension of these cells is determined by the operational flight altitude and the viewing angle of the onboard camera or sensor, ensuring that the covered area corresponds to the equipment's detection capacity. This discretizing transforms the physical map into a structure containing all possible actuation routes (E). After optimization and route acquisition for each UAV, the result can be visualized on the map (F).

Prior to the optimization phase, a network of feasible and safe flight corridors is generated to connect the identified POIs. The connections-graph is formed by the computer vision-based ground risk analysis algorithm introduced by \cite{p3}. Fundamentally, this paradigm allows the mission operator to explicitly tune of the cost function, setting a tunable tradeoff between maximizing path safety (risk avoidance) and minimizing travel distance (operational efficiency).

The UAV fleet is assumed to be homogeneous, with no provisions of energy harvesting or recharging. Energy is not only spent in displacing between points, but also during observations of the cells. The problem dictates that the energy consumed in a route cannot exceed the aircraft's maximum capacity.

A distinctive feature of this problem, in contrast to static approaches, is its random characteristic. The search environment changes as the operation progresses. When a UAV visits a cell and does not find the target, the probability of presence in that cell drops to near zero, which alters the global probability distribution of the map for the next iteration. This requires that route planning is not fixed from the start to the end of the mission, but rather adaptive.

Besides, the problem is treated with the flexibility of open routes. Operationally, this means UAVs are not required to spend energy returning to the original launch base at the end of each scanning cycle. They can terminate their current route at a strategic point in the field, maximizing useful search time in high-probability areas and minimizing unproductive displacements. Additional UAVs can also be easily added to the search at each iteration.

Route revisitation (overlap) becomes relevant whenever POI with high probability justify being visited more than once. These recurring visits can be performed by distinct UAVs, depending on routing optimization, imposed restrictions, and the determination of the operation manager, allowing for scanning by different search operators.

Therefore, the problem is summarized as deciding which sequence of cells each UAV must visit to maximize the accumulated probability of finding the victim, minimizing time and energy expenditure, while respecting fleet autonomy and the need for rapid coverage, which is vital for the targets' survival.

\subsection{Mathematical Model}

Our models are based on the OVRP with multiple depots \cite{p4}, which represents two separate optimization objectives that share a common set of constraints. Maximize total probability with a maximum time constraint, and minimize travel distance, while meeting a minimum probability collection.

Let $G = (V, A)$ be a directed graph, where the set of nodes $V = V_{depots} \cup V_{targets}$ is partitioned into a set of starting depots $V_{depots} = \{0, \dots, M-1\}$ and a set of target locations $V_{targets} = \{M, \dots, N\}$. The set of arcs is defined as $A = \{(i, j) : i, j \in V, i \neq j\}$. Let $K = \{0, \dots, M-1\}$ denote the set of available UAVs, where each UAV $k \in K$ is initially located at its corresponding depot $d_k \in V_{depots}$ (i.e., $d_k = k$). In addition, let $S \subseteq V_{targets}$, with $|S| \ge 2$, represent any proper subset of target nodes used for dynamic subtour elimination.

Each target node $i \in V_{targets}$ is associated with a probability (reward) $p_i$, while $t_{ij}$ denotes the time (or energy) cost required to traverse arc $(i, j)$. The parameter $E_k$ represents the maximum time (battery) capacity of UAV $k$, $P_{min}$ denotes the minimum required average probability, and $T_{max}$ defines the maximum allowable mission time. All parameters are normalized before solving, for numerical stability: each reward $p_i$ is written as a fraction of the total score available in the instance, and each travel cost $t_{ij}$ is divided by the largest arc weight. The binary variable $x_{ijk}$ is 1 when UAV $k$ flies arc $(i,j)$ and 0 otherwise; the integer variable $y_i$ counts how many times node $i$ is visited and is kept within $V_{min} \le y_i \le V_{max}$.

\paragraph{Model 1: Probability Maximization with Time Constraint.}

\begin{equation}
    \text{Maximize } Z_1 = \sum_{j \in V_{targets}} p_j y_j
    \label{eq:obj_z1}
\end{equation}

\begin{equation}
    \sum_{(i,j) \in A} t_{ij} x_{ijk} \leq T_{max}, \quad \forall k \in K
    \label{eq:time_limit}
\end{equation}

\begin{equation}
    \sum_{j \in V_{targets}} x_{kjk} \le 1, \quad \forall k \in K
    \label{eq:start_specific}
\end{equation}

\begin{equation}
    \sum_{i \in V} x_{ijk} = 0, \quad \forall j \in V_{depots}, \forall k \in K
    \label{eq:no_return}
\end{equation}

\begin{equation}
    \sum_{k \in K} \sum_{i \in V} x_{ijk} = y_j, \quad \forall j \in V_{targets}
    \label{eq:visit_link}
\end{equation}

\begin{equation}
    \sum_{i \in V} x_{ijk} \ge \sum_{i \in V} x_{jik}, \quad \forall j \in V_{targets}, \forall k \in K
    \label{eq:open_flow}
\end{equation}

\begin{equation} 
    \sum_{i \in V} \sum_{\substack{j \in V \\ j \neq i}} t_{ij} x_{ijk} \le E_k, \quad \forall k \in K
    \label{eq:energy}
\end{equation}

\begin{equation}
    \sum_{i \in S} \sum_{j \in S} x_{ijk} \le |S|(|S|-1) \cdot
    \sum_{i \in V \setminus S} \sum_{j \in S} x_{ijk},
    \quad \forall k \in K, \forall S \subseteq V_{targets}, |S| \ge 2
    \label{eq:sec_lazy}
\end{equation}

\begin{equation} 
    x_{ijk} \in \{0, 1\}, \quad \forall (i, j) \in A, \forall k \in K 
    \label{eq:dom_x}
\end{equation}

\begin{equation} 
    y_j \in \{V_{min}, \dots, V_{max}\}, \quad \forall j \in V_{targets} 
    \label{eq:dom_y}
\end{equation}

Model~1 aims to maximize the total collected probability by selecting and routing UAVs to visit target nodes, subject to a strict upper bound on the mission duration of each UAV. The objective function~\eqref{eq:obj_z1} maximizes the sum of the probabilities associated with the visited targets. Constraint~\eqref{eq:time_limit} limits the total travel time of each UAV, ensuring feasibility with respect to the imposed mission horizon. Constraints~\eqref{eq:start_specific} and~\eqref{eq:no_return} enforce that each UAV departs exclusively from its corresponding depot and that no UAV enters any depot node. Constraint~\eqref{eq:visit_link} links the routing decisions to the visit counter, guaranteeing that a target is considered visited if and only if it is reached by at least one UAV. The OVRP structure is imposed by constraint~\eqref{eq:open_flow}. Constraint~\eqref{eq:energy} limits the maximum energy consumption of each UAV, while constraint~\eqref{eq:sec_lazy} corresponds to dynamically generated subtour elimination constraints that prevent the formation of isolated cycles over subsets of target nodes. Finally, constraints~\eqref{eq:dom_x} and~\eqref{eq:dom_y} establish the domain of the decision variables, restricting routing decisions to binary values and enforcing integer bounds on the visit counters.

\paragraph{Model 2: Distance Minimization with Probability Requirement.}

\begin{equation}
    \text{Minimize } Z_2 = \sum_{k \in K} \sum_{(i,j) \in A} t_{ij} x_{ijk}
    \label{eq:obj_z2}
\end{equation}

\begin{equation}
    \sum_{j \in V_{targets}} p_j y_j \ge P_{min}
    \label{eq:min_prob}
\end{equation}

All constraints \eqref{eq:start_specific}--\eqref{eq:dom_y} remain unchanged.

Model~2 looks at the same problem from the other side: now it is the travel distance that is minimized. Objective~\eqref{eq:obj_z2} takes the place of~\eqref{eq:obj_z1}, the time limit~\eqref{eq:time_limit} is dropped, and constraint~\eqref{eq:min_prob} asks instead for a minimum amount of collected probability. Everything else is carried over from Model~1 unchanged, so the two models share one routing structure and differ only in what they optimize and what they demand.

\section{The RKO-based Solution Framework}
\label{sec:method}

Our implementation runs several metaheuristics in parallel, coordinated by the RKO. When one of them stalls in a local optimum, as single-trajectory methods often do, it restarts from a solution another method has placed in the shared elite pool. The probabilistic landscape of a SAR mission is full of such optima; every cluster of high-probability cells is one. A trajectory method tends to settle in the nearest cluster. A population keeps enough diversity to work on several distant clusters at the same time, which RKO achieves by mapping continuous random keys to valid routes.

\subsection{Decoder Architecture}

Much of the work in this paper is in the decoder, which has to cope with two objectives and a dense set of constraints at once. A solution is a vector of random keys. The decoding process maps this continuous vector into a discrete, feasible routing plan through a two-stage process.

\subsubsection{Stage 1: Route construction through delimiter based sorting }

The first part of the chromosome, whose length equals $|V_{depots}| + |V_{targets}|$, decides the order and grouping. The set $V = V_{depots} \cup V_{targets}$ contains every UAV depot and every possible target. The decoder generates a permutation  by sorting the indices based on the ascending order of their corresponding random keys. We employ a delimiter-based decoding strategy, which is particularly effective for multi-vehicle problems. In this paradigm, the depot nodes $g \in V_{depots}$ act as delimiters within the permutation $\pi$, partitioning the sequence of customer nodes into distinct clusters. 

\begin{figure}[htbp] 
\centering 
\includegraphics[width=\textwidth]{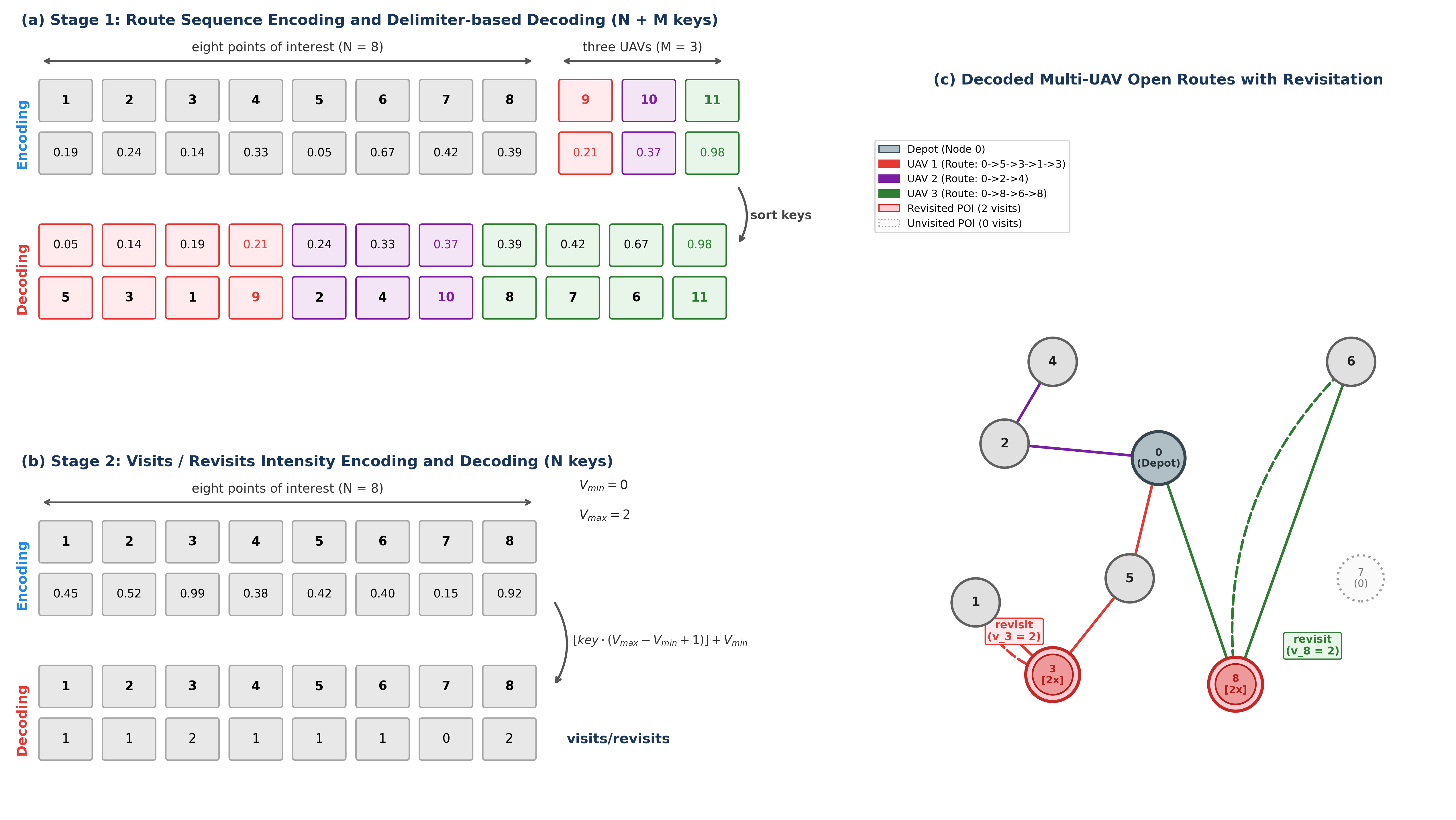} 
\caption{Two-stage decoding of a random-key vector into a multi-UAV plan, for $N = 8$ points of interest and $M = 3$ UAVs. (a) Stage 1: the first $N+M$ keys are sorted; the depot keys (9, 10 and 11) act as delimiters and split the sorted sequence into one route per UAV. (b) Stage 2: the remaining $N$ keys are mapped to visit counts in $[V_{min}, V_{max}]$, here $[0,2]$. (c) Resulting open routes. Node 7 receives zero visits and is left out; nodes 3 and 8 are visited twice, the second visit inserted at the cheapest position in the route (dashed arcs).}
\label{fig:rko_decoding}
\end{figure}

The mechanism lets the random keys fix at the same time which UAV serves which customer and the sequence in which each UAV will visit its stops, as shown in Fig.~\ref{fig:rko_decoding}.

\subsubsection{Stage 2: Insertion for multi visit intensity }
Standard VRP visit each stop once, our model visits a stop as often as the mission needs but never more than $V_{max}$ times. The second part of the keys sets this rule with $V = V_{targets}$. For every target node $i$ the key stores a random key $\rho_i \in [0, 1]$ this key fixes the number of visits and extra visits, written $\nu_i$. The key is a real number, we slice it into an integer that lies between the least and the greatest allowed visits, $V_{min}$ and $V_{max}$. Consequently, the total number of planned visits to node $i$ becomes $V_{min} + \nu_i$. Every extra stop is pushed into the tour at the spot that adds the least extra time, with $visits=0$ the stop not enter in route.

\subsubsection{Constraint Handling and Objective Evaluation}

The decoder treats constraints such as battery capacity as soft constraints during the evaluation phase. If a constructed route exceeds the energy limit (~\ref{eq:energy}) or application of penalties based (~\ref{eq:time_limit}) and (~\ref{eq:min_prob}), a penalty proportional to the violation is added to the objective function~\eqref{eq:penalized}, where $\lambda$ is a large penalty factor. This allows the evolutionary algorithm to traverse infeasible regions of the search space to reach feasible optima.

\begin{equation} 
f_{penalized} = f_{obj} + \lambda \cdot \sum_{k \in K} \max(0, \text{Cost}_k - E_k)
\label{eq:penalized}
\end{equation}

\subsection{Dynamic Search Area Characterization}

\begin{figure}[htbp]
    \centering
    \includegraphics[width=\textwidth]{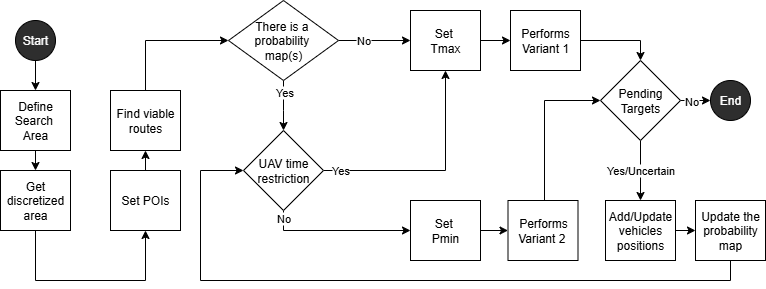}
    \caption{Dynamic search loop. The area is discretized and the feasible routes are built once, before the first flight. At each iteration the planner chooses between Model~1 (time limit $T_{max}$) and Model~2 (coverage floor $P_{min}$) depending on the available information and on the operational constraints, solves it with the RKO, and, while targets remain pending, updates the vehicle positions and the probability map before planning the next segment.}
    \label{fig:methodology_flow}
\end{figure}

Unlike static routing methods that assume a stationary environment, the proposed framework dynamically adapts search trajectories. Each revision reshapes the goal, balancing what is known in a probability map against the limits in minutes left in the air against ground that must be seen. Figure ~\ref{fig:methodology_flow} shows the loop. 

The loop starts on the ground: the region is cut into grid cells, the POIs are picked out and the feasible flight corridors are built between them. The planner then reads the mission parameters and chooses the formulation. Model~1 is used when the UAVs fly under a hard time limit or when no prior map is available; Model~2 when a prior map exists and the binding requirement is a coverage floor.

These mathematical formulations serve strictly to define the objectives and operational constraints; they do not participate directly in the execution of the search architecture. Instead, the proposed RKO acts as the core solver tool, providing a unified framework capable of solving both Model~1 and Model~2 seamlessly. By encoding route decisions into continuous random key vectors, RKO decouples solution representation from problem-specific constraints. As updated probability maps or operational telemetry are received during the mission, RKO rapidly re-optimizes the trajectories, ensuring agile adaptation within the dynamic search loop.

\section{Computational Experiments}
\label{sec:results}

We tested the framework on synthetic missions that mirror the randomness of real SAR operations and that scale in difficulty.

\subsection{Instance Generation and Probability Landscape}

The probability landscape is a Gaussian mixture. Its cluster centers $\mathcal{C} = \{c_1, c_2, \dots, c_K\}$ sit where a victim is most likely to be, and each center comes from one of three sources:
\begin{itemize}
    \item \textbf{Historical data:} the places where incidents come back year after year.
    \item \textbf{Expert advice:} slopes, riverbanks and the other hazards the specialists point to.
    \item \textbf{Live intelligence:} the latest sighting, or the last known position.
\end{itemize}
Figure~\ref{mapa}(D) shows the surface that results; the second stage turns it into a priority, or profit, for each node.

We do not evaluate the full mixture at every node. It would cost more than it is worth, and it blurs the borders between regions. We use instead a localized $k$-Nearest Neighbor ($k$-NN) kernel. For a generated node $i$ at location $(x_i, y_i)$, we identify the subset $\mathcal{N}_i \subset \mathcal{C}$ containing the $N$ closest cluster centers (where $|\mathcal{N}_i| = N$). The base spatial probability $P_{spatial}(i)$ is the sum of the Gaussian influences from these local neighbors.

The $P_i$ final score is modulated by the environmental context extracted from the underlying map zones (e.g., Red, Orange, Green zones in Figure~\ref{mapa}). Let $W_{zone}(i)$ be the weight multiplier associated with the terrain color at coordinates $(x_i, y_i)$. The final priority score for node $i$ is defined as:

\begin{equation}
    P_i = \lfloor P_{spatial}(i) \cdot W_{zone}(i) \rfloor
\end{equation}

\subsection{Computational Results}\label{sec:comp_results}

In this section, the performance of the proposed RKO framework is evaluate against a exact solver, Gurobi Optimizer version 12.0.3 in Ubuntu 22.04.4 LTS, CPU Intel(R) Xeon(R) 2.20GHz. Table (\ref{tab:results}) summarizes the comparison across three instances groups ranging from 3 UAVs, 1000 nodes, $|V|=10$ to $|V|=100$ POIs, encompassing both the minimizing distance subject to $P_{min}$ and maximizing probability subject to $T_{max}$ models. Following \cite{i6}, who set the time limit proportionally to instance size (one second per node in the GQAP and THLP experiments), we adopt a single wall-clock limit of 120 s, which exceeds one second per node for all instances ($\geq 1.2 s/node$, $\geq 0.6 s/key$) and is also given to Gurobi, and includes the solution GAP, and RKO includes the sample standard deviation based on 10 runs of the framework.

\begin{table}[ht]
\centering
\caption{Comparative performance of the exact method (Gurobi) and the RKO metaheuristic across instances, objective functions, and visit constraints}
\label{tab:results}
\begin{tabular}{ccccccccccc} 
\toprule
& & & & & \multicolumn{3}{c}{\textbf{Exact method (Gurobi)}} & \multicolumn{3}{c}{\textbf{RKO}} \\
\cmidrule(lr){5-7} \cmidrule(lr){8-10}
& ID & Visits & Objective & Const. & \makecell{Obj. \\Value} & \makecell{Time \\(s)} & \makecell{Gap \\(\%)} & \makecell{Obj. \\Value} & \makecell{Time \\(s)} & \makecell{Dev. \\(\%)} \\ 
\midrule
\multirow{8}{*}{\rotatebox{90}{Instance 1 $|V|=10$}} 
 & i1 & [1,3] & $max\ Z_1$ & $40$ & \textbf{0.750} & 0.26 & 0.00 & \textbf{0.750} & 2.058 & 0.083 \\
 & i2 & [0,3] & $max\ Z_1$ & $40$ & \textbf{0.770} & 3.22 & 0.00 & \textbf{0.770} & 30.243 & 0.127 \\
 & i3 & [1,3] & $max\ Z_1$ & $60$ & \textbf{0.785} & 0.02 & 0.00 & \textbf{0.785} & 6.449 & 0.319 \\
 & i4 & [0,3] & $max\ Z_1$ & $60$ & \textbf{0.815} & 5.11 & 0.00 & \textbf{0.815} & 28.166 & 0.455 \\
 & i5 & [1,3] & $min\ Z_2$ & $0.8$ & \textbf{1.716 }& 0.05 & 0.00 & \textbf{1.716} & 4.023 & 0.232 \\
 & i6 & [0,3] & $min\ Z_2$ & $0.8$ & \textbf{1.006} & 0.01 & 0.00 & \textbf{1.006} & 0.085 & 0.056 \\
 & i7 & [1,3] & $min\ Z_2$ & $0.3$ & \textbf{3.825} & 4.89 & 0.00 & \textbf{3.825} & 27.331 & 0.032 \\
 & i8 & [0,3] & $min\ Z_2$ & $0.3$ & \textbf{0.705} & 0.00 & 0.00 & \textbf{0.705} & 0.087 & 0.047 \\
\midrule
\multirow{8}{*}{\rotatebox{90}{Instance 2 $|V|=50$}} 
  & i9 & [1,3] & $max\ Z_1$ & $200$ & \textbf{0.990} & 120.00 & 2.07 & 0.963 & 105.52 & 0.034 \\
  & i10 & [0,3] & $max\ Z_1$ & $200$ & 0.941 & 120.00 & 6.22 & \textbf{0.970} & 107.65 & 0.024 \\
  & i11 & [1,3] & $max\ Z_1$ & $150$ & - & - & - & \textbf{0.855} & 109.18 & 0.029 \\
  & i12 & [0,3] & $max\ Z_1$ & $150$ & 0.614 & 120.00 & 60.3 & \textbf{0.909} & 112.31 & 0.030 \\
  & i13 & [1,3] & $min\ Z_2$ & $0.8$ & 8.030 & 120.00 & 1.91 & \textbf{7.846} & 108.45 & 0.085 \\
  & i14 & [0,3] & $min\ Z_2$ & $0.8$ & 6.455 & 120.00 & 37.2 & \textbf{5.946} & 101.76 & 0.627 \\
  & i15 & [1,3] & $min\ Z_2$ & $0.3$ & \textbf{5.117} & 14.61 & 0.00 & 5.260 & 39.823 & 0.297 \\
  & i16 & [0,3] & $min\ Z_2$ & $0.3$ & \textbf{1.418} & 3.72 & 0.00 & 1.529 & 47.852 & 0.028 \\
  \midrule
\multirow{8}{*}{\rotatebox{90}{Instance 3 $|V|=100$}} 
  & i17 & [1,3] & $max\ Z_1$ & $350$ & 0.950 & 120.00 & 5.20 & \textbf{0.956} & 120.00 & 0.032 \\
  & i18 & [0,3] & $max\ Z_1$ & $350$ & \textbf{0.960} & 53.66 & 0.00 & \textbf{0.960} & 119.96 & 0.029 \\
  & i19 & [1,3] & $max\ Z_1$ & $500$ & \textbf{0.992} & 0.40 & 0.00 & \textbf{0.992} & 120.00 & 0.026 \\
  & i20 & [0,3] & $max\ Z_1$ & $500$ & \textbf{0.993} & 15.33 & 0.00 & \textbf{0.993} & 120.00 & 0.016 \\
  & i21 & [1,3] & $min\ Z_2$ & $0.8$ & 58.125 & 120.00 & 85.2 & \textbf{14.946} & 120.00 & 0.114 \\
  & i22 & [0,3] & $min\ Z_2$ & $0.8$ & - & - & - & \textbf{13.344} & 120.01 & 0.453 \\
  & i23 & [1,3] & $min\ Z_2$ & $0.3$ & \textbf{6.609} & 101.17 & 0.00 & 6.639 & 114.03 & 0.510 \\
  & i24 & [0,3] & $min\ Z_2$ & $0.3$ & \textbf{3.116} & 120.00 & 34.9 & 3.258 & 91.989 & 0.384 \\
  \bottomrule
\end{tabular}
\end{table}

For small-scale instances, the exact method efficiently proves optimality. In these scenarios, the RKO framework manifests  high  performance, achieving optimal solutions. Although the heuristic runtime is higher than the exact solver for these trivial instances due to the initialization overhead of the evolutionary framework, the absolute difference is operationally insignificant. The picture changes as the instances grow. In the medium group the exact solver starts to pay the exponential price that NP-hard routing problems charge: it still proves optimality for the two $Z_2$ cases with the low requirement, but on the others it runs to the 120-second limit with the gap still open, and in one configuration it returns no feasible solution at all.

The largest group shows where the exact approach stops being usable for this class of problem. In half of its configurations Gurobi proves optimality; in the other half it either stops at the time limit with an open gap or finds no feasible solution, and where the gap stays open the RKO value is ahead of the incumbent in two of the three cases, by a wide margin in one of them.

Delivering a good plan inside a two-minute limit is what makes the RKO fit for field use, where a workable plan now is worth more than a provably best one later. In short, the exact model gives the reference values on the small cases, and the RKO gives the robustness and speed the large ones demand, keeping exploration and exploitation in balance along the way.

\subsection{Comparative Analysis with Metaheuristics}

To see how the RKO holds up under the same computational budget, we ran it against three single-trajectory metaheuristics in common use for UAV routing: LNS, ILS and VNS. Each was run 10 times on each instance, and every run was cut off at a hard limit of 120 seconds Table (~\ref{tab:results2}). The column marked \textit{Obj.} states the mean objective value found, the column T(s) states the average execution time, in seconds, seen across the attempts. The instances centre on larger cases where the search space grows markedly.

\begin{table}[ht]
\centering
\caption{Comparison between RKO and other metaheuristics ($|V| = 100$). Obj. Value: average over 10 runs. All methods ran with the same 120\,s wall-clock limit; RKO runs its 10 metaheuristics in parallel (one thread each), while LNS, ILS and VNS run alone in a single thread.}
\label{tab:results2}
\begin{tabular}{ccccccccccccc}
\toprule
& \multicolumn{3}{c}{\textbf{RKO}} & \multicolumn{3}{c}{\textbf{LNS}} & \multicolumn{3}{c}{\textbf{ILS}} & \multicolumn{3}{c}{\textbf{VNS}} \\
\cmidrule(lr){2-4} \cmidrule(lr){5-7} \cmidrule(lr){8-10} \cmidrule(lr){11-13}
ID & \makecell{Obj. \\Value} & \makecell{Time \\(s)} & \makecell{Dev. \\(\%)} & \makecell{Obj. \\Value} & \makecell{Time \\(s)} & \makecell{Dev. \\(\%)} & \makecell{Obj. \\Value} & \makecell{Time \\(s)} & \makecell{Dev. \\(\%)} & \makecell{Obj. \\Value} & \makecell{Time \\(s)} & \makecell{Dev. \\(\%)} \\ 
\midrule
 i17 & \textbf{0.956} & 115.3 & 1.24 & 0.791 & 118.3 & 2.76 & 0.861 & 116.9 & 2.56 & 0.872 & 114.7 & 2.42 \\
 i18 & \textbf{0.960} & 116.4 & 1.27 & 0.837 & 108.6 & 1.88 & 0.891 & 117.5 & 1.22 & 0.896 & 112.1 & 0.59 \\
 
 i19 & \textbf{0.992} & 80.0 & 0.09 & 0.901 & 111.2 & 1.09 & 0.974 & 115.5 & 0.73 & 0.976 & 111.3 & 0.93 \\
 i20 & \textbf{0.993} & 105.8 & 0.40 & 0.896 & 113.0 & 0.92 & 0.950 & 116.2 & 0.78 & 0.953 & 110.6 & 0.90 \\
 
 i21 & \textbf{14.946} & 114.6 & 6.48 & 19.402 & 118.7 & 5.21 & 16.134 & 115.7 & 7.46 & 15.124 & 114.0 & 7.63 \\
 i22 & \textbf{13.344} & 114.2 & 5.36 & 18.258 & 118.2 & 5.63 & 14.666 & 116.5 & 4.15 & 14.131 & 116.9 & 6.92 \\ 
 
i23 & \textbf{6.639} & 78.0 & 3.32 & 8.841 & 112.7 & 5.00 & 8.056 & 97.3 & 3.54 & 8.088 & 107.2 & 3.21 \\
 i24 & \textbf{3.258} & 82.8 & 9.59 & 3.867 & 87.0 & 14.09 & 3.509 & 102.2 & 6.72 & \textbf{3.258} & 89.8 & 7.67 \\
 
 \bottomrule
\end{tabular}
\end{table}

The RKO framework achieved the best average objective value in all scenarios. A distinct behavior was observed only in a few instances, where all methods converged to the same objective value. In the more constrained scenarios, however, the differences between the methods become more pronounced, with methods such as LNS reaching worse solutions than RKO. We attribute this advantage to the collaboration among RKO’s parallel metaheuristics: by sharing high-quality solutions through the elite pool, each metaheuristic can continue its search from promising regions found by the others, allowing the framework to reach better routing configurations than any single method operating alone. The superiority of the RKO framework is most evident in the minimization instances, where the goal is to satisfy the coverage requirements at minimal cost.

\subsection{Degradation Under Noisy and Biased Prior Maps}

In real-world SAR missions, initial probability maps are rarely perfect. We tested the robustness of the framework by introducing controlled imperfections to the prior map: random noise ($\sigma$) and systematic bias ($b$). We compared these against an exact map and against a ``blind'' planner, which flies on a uniform map and so with no prior at all Table (~\ref{tab:map_degradation}).

\begin{table}[ht]
\centering
\caption{Framework Performance Under Varying Degrees of Map Degradation}
\label{tab:map_degradation}
\begin{tabular}{lc}
\toprule
\textbf{Planner Map Quality} & \textbf{Scan Success (\%)} \\
\midrule
Exact & \textbf{52.46} \\
Noise $\sigma = 0.5$ & 48.39 \\
Bias $b = 0.5$ & 44.50 \\
Noise $\sigma = 1.0$ & 40.97 \\
\textbf{Blind (Uniform Map)} & \textbf{37.60} \\
Bias $b = 1.0$ & \textbf{20.23} \\
\bottomrule
\end{tabular}
\end{table}

Noise is absorbed well. Even at $\sigma = 1.0$ the planner still finds 40.97\% of the victims, ahead of the blind run. Noise of this kind is scattered rather than aligned, so the zones that really carry high probability still stand out from it, and the routes still go there.

Bias is another matter. At $b = 1.0$ the score falls to 20.23\%, little more than half of what the blind planner gets. For a SAR team the lesson is uncomfortable but plain: intelligence that is wrong in a consistent direction does more harm than admitting there is none. If the intelligence is highly questionable, it is mathematically safer to assume a uniform probability map.

\subsection{Comparative Analysis with Myopic Replanning}

We evaluated whether the global optimization provided by the RKO framework outperforms a simpler, myopic replanning strategy. Five operational baselines were defined by varying three things independently: whether the planner replans at all, whether it holds a prior probability map, and which engine makes the decision. Every arm flies the same scenarios under common random numbers.

The myopic baseline is a standard greedy construction for prize-collecting routing. At each step it advances the aircraft with the smallest clock and, for that aircraft, selects the candidate with the best prize density $\tau_{ij} = p_j / (t_{ij} + T_O)$, subject to the same feasibility rules the global planner obeys: the visit cap $V_{max}$, the minimum $V_{min}$, the battery with its return reserve, and the mission horizon. It is myopic in lookahead but maximally reactive in state, and it costs no solver time.

\begin{table}[ht]
\centering
\caption{Performance comparison across replanning and search strategies. All arms use the linear objective of Equation ~\ref{eq:obj_z1} and obey the same constraints. Mean over two instances and five Monte Carlo replications.}
\label{tab:myopic_comp}
\begin{tabular}{lcccrr}
\toprule
\textbf{Strategy} & \textbf{Replans} & \textbf{Map} & \textbf{Decision engine} & \textbf{\%scan} & \textbf{Visits} \\
\midrule
RKO Static   & No  & Exact   & Global search & 58.91 & 56.8 \\
RKO Complete & Yes & Exact   & Global search & 56.01 & 73.5 \\
Myopic       & Yes & Exact   & Greedy        & 49.45 & 105.5 \\
Blind-Static & No  & Uniform & Global search & 48.79 & 67.2 \\
Blind        & Yes & Uniform & Global search & 43.74 & 92.3 \\
\bottomrule
\end{tabular}
\end{table}

Table (~\ref{tab:pairwise_myopic}) reports the paired comparisons. The global search beats the greedy engine, and by a wide margin: 56.01 against 49.45, with the greedy spending 44\% more observations to get there. The prior map also pays for itself, and removing both the map and the replanning drops the method to the floor of the comparison. These two results support the two design choices the framework rests on.

\begin{table}[ht]
\centering
\caption{Paired comparisons against the Complete method (Wilcoxon signed-rank,
Holm-corrected, ten instance--seed cells).}
\label{tab:pairwise_myopic}
\begin{tabular}{lccccl}
\toprule
\textbf{Comparison} & \textbf{Win} & \textbf{Tie} & \textbf{Loss} & \textbf{$p$ (Holm)} & \textbf{Reading} \\
\midrule
RKO Complete vs Myopic       & 10 & 0 & 0 & 0.0078 & global search pays \\
RKO Complete vs Blind        &  9 & 0 & 1 & 0.0293 & the prior map pays \\
RKO Complete vs Blind-Static &  8 & 0 & 2 & 0.0293 & higher scan success \\
RKO Complete vs RKO Static       &  4 & 1 & 5 & 0.4258 & no measurable gain \\
\bottomrule
\end{tabular}
\end{table}

Replanning on its own shows no measurable benefit: the Static arm sits slightly ahead of the Complete one and the paired test cannot separate them. We report this rather than set it aside, under a fixed total budget, every extra replan buys new information at the cost of a thinner solver slice, and in this configuration the two effects cancel.

\subsection{Comparative Analysis with Iteration Modes}

We now test how the Dynamic Search Area Characterization method affects the mission. Earlier tests used fixed conditions; here the planner must revise the route again and again while the flight proceeds. Table (~\ref{tab:iter_results}) lists the results for each mode.

A strategy fixes two things at once: which objective is active ($Z_1$ maximises collected probability, $Z_2$ minimises travel) and how the constraint moves along the mission. The constraint is written as a fraction: under $Z_1$ it is the share of the battery $E_k$ the aircraft may spend on one route, and under $Z_2$ it is the coverage floor $P_{min}$. A single number fixes a hard bound for the whole flight; an arrow gives a bound that changes after every replan, either tightening or loosening step by step. The column \%scan reports the share of victims actually found, \emph{Vis.} sums the observations made at POIs, and \emph{Rep.} counts how many times the planner rebuilt the route.

Groups A and B are the control. They are the only strategies a static planner could also execute, since neither the objective nor the limit ever changes. Everything from group C onward requires replanning, so the distance between those groups and the control is what iteration buys. The same instance is flown by one, two and three UAVs, so the table answers two questions at once: whether iteration helps, and whether what it adds is still there once the fleet grows.

\begin{table}[ht]
\centering
\caption{Performance of dynamic re-optimization strategies. Each strategy fixes how the objective and the constraint move along the mission; the constraint is a fraction of $E_k$ under $Z_1$ and a coverage floor under $Z_2$. Groups A and B are the control: a static planner could execute them, because nothing changes during the flight. The instance has 10 POIs, $[0,2]$ visits and a horizon of three trips.}
\label{tab:iter_results}
{\footnotesize
\begin{tabular}{clccccccccc}
\toprule
& & \multicolumn{3}{c}{1 UAV} & \multicolumn{3}{c}{2 UAVs} & \multicolumn{3}{c}{3 UAVs} \\
\cmidrule(lr){3-5} \cmidrule(lr){6-8} \cmidrule(lr){9-11}
Group & Strategy & \%scan & Vis. & Rep. & \%scan & Vis. & Rep. & \%scan & Vis. & Rep. \\
\midrule
\multirow{4}{*}{\rotatebox{90}{\textbf{$Z_1$ fixed}}} 
 & $Z_1$ fixed 0.4 & 63.86 & 13 & 3 & 40.96 & 24 & 10 & 59.04 & 36 & 12 \\
 & $Z_1$ fixed 0.6 & 75.90 & 39 & 9 & 69.88 & 32 & 9 & 68.67 & 27 & 7 \\
 & $Z_1$ fixed 0.8 & 79.52 & 36 & 6 & \textbf{75.90} & 38 & 9 & 78.31 & 36 & 9 \\
 & $Z_1$ fixed 1.0 & 79.52 & 36 & 6 & 73.49 & 27 & 6 & \textbf{93.98} & 33 & 9 \\
\midrule
\multirow{5}{*}{\rotatebox{90}{\textbf{$Z_2$ fixed}}} 
 & $Z_2$ fixed 0.1 & 4.82 & 126 & 64 & 21.69 & 98 & 64 & 42.17 & 100 & 64 \\
 & $Z_2$ fixed 0.3 & 14.46 & 126 & 64 & 39.76 & 68 & 35 & 51.81 & 64 & 34 \\
 & $Z_2$ fixed 0.5 & 63.86 & 57 & 19 & 38.55 & 36 & 11 & 51.81 & 52 & 16 \\
 & $Z_2$ fixed 0.7 & 59.04 & 57 & 14 & 62.65 & 36 & 11 & 59.04 & 29 & 10 \\
 & $Z_2$ fixed 0.9 & 65.06 & 38 & 10 & 57.83 & 37 & 10 & 61.45 & 36 & 10 \\
\midrule
\multirow{4}{*}{\rotatebox{90}{\textbf{$Z_1$ ramp}}} 
 & $Z_1$ 1.0$\to$0.2 (tighten) & 69.88 & 12 & 3 & 72.29 & 24 & 6 & 78.31 & 38 & 10 \\
 & $Z_1$ 0.8$\to$0.4 (tighten) & 84.34 & 25 & 5 & 73.49 & 32 & 10 & 75.90 & 30 & 8 \\
 & $Z_1$ 0.2$\to$1.0 (loosen) & 79.52 & 34 & 7 & 72.29 & 32 & 10 & 78.31 & 44 & 12 \\
 & $Z_1$ 0.4$\to$0.8 (loosen) & 86.75 & 37 & 7 & 74.70 & 35 & 11 & 84.34 & 28 & 8 \\
\midrule
\multirow{4}{*}{\rotatebox{90}{\textbf{$Z_2$ ramp}}} 
 & $Z_2$ 0.9$\to$0.1 (loosen) & 60.24 & 122 & 64 & 57.83 & 98 & 64 & 72.29 & 54 & 25 \\
 & $Z_2$ 0.7$\to$0.3 (loosen) & 34.94 & 131 & 64 & 61.45 & 43 & 19 & 59.04 & 46 & 21 \\
 & $Z_2$ 0.1$\to$0.9 (tighten) & 77.11 & 39 & 11 & 57.83 & 43 & 14 & 72.29 & 42 & 14 \\
 & $Z_2$ 0.3$\to$0.7 (tighten) & 59.04 & 59 & 16 & 60.24 & 22 & 9 & 75.90 & 35 & 15 \\
\midrule
\multirow{12}{*}{\rotatebox{90}{\textbf{Two-phase}}} 
 & $Z_1\to Z_2$ @25\% (0.8, 0.3) & 72.29 & 120 & 55 & 60.24 & 70 & 33 & 77.11 & 59 & 26 \\
 & $Z_1\to Z_2$ @25\% (0.5, 0.6) & 51.81 & 24 & 6 & 61.45 & 55 & 15 & 67.47 & 57 & 15 \\
 & $Z_2\to Z_1$ @25\% (0.3, 0.8) & 78.31 & 36 & 7 & \textbf{75.90} & 36 & 10 & 92.77 & 50 & 15 \\
 & $Z_2\to Z_1$ @25\% (0.6, 0.5) & 86.75 & 43 & 10 & 73.49 & 37 & 9 & 59.04 & 22 & 7 \\
 & $Z_1\to Z_2$ @50\% (0.8, 0.3) & 71.08 & 120 & 56 & 62.65 & 66 & 25 & 78.31 & 59 & 24 \\
 & $Z_1\to Z_2$ @50\% (0.5, 0.6) & \textbf{91.57} & 56 & 13 & 61.45 & 52 & 14 & 72.29 & 20 & 6 \\
 & $Z_2\to Z_1$ @50\% (0.3, 0.8) & 78.31 & 37 & 8 & \textbf{75.90} & 36 & 11 & 77.11 & 32 & 12 \\
 & $Z_2\to Z_1$ @50\% (0.6, 0.5) & 69.88 & 16 & 5 & 61.45 & 34 & 9 & 59.04 & 32 & 10 \\
 & $Z_1\to Z_2$ @75\% (0.8, 0.3) & 77.11 & 88 & 37 & 73.49 & 54 & 22 & 91.57 & 52 & 21 \\
 & $Z_1\to Z_2$ @75\% (0.5, 0.6) & 73.49 & 59 & 15 & 73.49 & 28 & 7 & 55.42 & 12 & 4 \\
 & $Z_2\to Z_1$ @75\% (0.3, 0.8) & 79.52 & 40 & 8 & \textbf{75.90} & 40 & 14 & 85.54 & 42 & 16 \\
 & $Z_2\to Z_1$ @75\% (0.6, 0.5) & 78.31 & 50 & 12 & 61.45 & 39 & 11 & 68.67 & 28 & 10 \\
\midrule
\multirow{8}{*}{\rotatebox{90}{\textbf{Alternating}}} 
 & $Z_1|Z_2$ alternat. (0.8/0.3) & 77.11 & 88 & 37 & 73.49 & 74 & 34 & 71.08 & 81 & 37 \\
 & $Z_1|Z_2$ alternat. (0.5/0.6) & 55.42 & 65 & 15 & 61.45 & 61 & 18 & 55.42 & 12 & 4 \\
 & $Z_2|Z_1$ alternat. (0.3/0.8) & 78.31 & 36 & 7 & 73.49 & 29 & 9 & 92.77 & 40 & 12 \\
 & $Z_2|Z_1$ alternat. (0.6/0.5) & 86.75 & 43 & 10 & 61.45 & 39 & 11 & 57.83 & 22 & 6 \\
 & $Z_1\to Z_2\to Z_1$ (0.8/0.3) & 79.52 & 36 & 6 & 73.49 & 33 & 10 & 91.57 & 38 & 12 \\
 & $Z_1\to Z_2\to Z_1$ (0.5/0.6) & 80.72 & 50 & 11 & 73.49 & 31 & 8 & 54.22 & 12 & 4 \\
 & $Z_2\to Z_1\to Z_2$ (0.3/0.8) & 61.45 & 132 & 64 & 60.24 & 69 & 33 & 77.11 & 60 & 26 \\
 & $Z_2\to Z_1\to Z_2$ (0.6/0.5) & 72.29 & 54 & 14 & 61.45 & 35 & 9 & 59.04 & 24 & 8 \\
\midrule
\multirow{4}{*}{\rotatebox{90}{\textbf{Ramp}}} 
 & $Z_1\to Z_2$, tightening ramp & 86.75 & 138 & 64 & 72.29 & 24 & 5 & 75.90 & 65 & 27 \\
 & $Z_1\to Z_2$, loosening ramp & 67.47 & 77 & 21 & 62.65 & 61 & 20 & 69.88 & 51 & 17 \\
 & $Z_2\to Z_1$, tightening ramp & 72.29 & 26 & 6 & 61.45 & 40 & 13 & 67.47 & 29 & 9 \\
 & $Z_2\to Z_1$, loosening ramp & 77.11 & 39 & 8 & 40.96 & 8 & 4 & 67.47 & 37 & 11 \\
\bottomrule
\end{tabular}
}
\end{table}

The first group runs on $Z_1$ and collects as much probability as it can inside a strict travel allowance. The link between spare budget and mission success is not a straight line. Raising the allowance lifts the score steeply at first, then the curve goes flat: with one UAV, 0.8 and 1.0 both end at 79.52, because by then the battery is no longer the binding limit and handing the aircraft more of it changes nothing.

The second group runs on $Z_2$ and flies until the coverage floor $P_{min}$ is met. A low floor turns out to be harmful rather than lenient. With $P_{min} = 0.1$ the planner does the least work that meets the target, the plan is used up almost at once and the loop has to build another; that configuration hits the ceiling of 64 replans with every fleet size and still posts the worst scores of the grid, after a hundred or so observations. The aircraft spends the mission circling cheap ground. A higher floor forces longer flights, and more useful ones.

Groups C to G let the limit move, the objective switch, or both, and what they are worth depends sharply on how many aircraft are in the air Table (~\ref{tab:fleet_gain}).

\clearpage
\begin{table}[!ht]
\centering
\caption{Best fixed strategy (groups A--B) against the best adaptive one (groups C--G), by fleet size. Values are \%scan; the gain is the difference in percentage points.}
\label{tab:fleet_gain}
\begin{tabular}{lccc}
\toprule
Fleet & Best fixed (A--B) & Best adaptive (C--G) & Gain \\
\midrule
1 UAV  & 79.52 & \textbf{91.57} & $+12.05$ \\
2 UAVs & \textbf{75.90} & \textbf{75.90} & $\pm 0.00$ \\
3 UAVs & \textbf{93.98} & 92.77 & $-1.20$ \\
\bottomrule
\end{tabular}
\end{table}

With a single aircraft the schedule pays its way: switching from $Z_1$ to $Z_2$ at the midpoint of the mission reaches 91.57, twelve points above the best fixed setting. While the budget is generous the aircraft goes after the promising cells; once most of the map has been cleared it switches to spending less, and stays useful for longer.

That advantage does not survive a larger fleet. With two UAVs the best adaptive strategy only matches the best fixed one, and with three the best plan of the whole grid is a fixed setting $Z_1$ held at its loosest bound for the entire mission, at 93.98. The extra aircraft supply the same flexibility the schedule was supplying before: with several routes in the air the planner can pursue different aims at once and no longer needs to alternate between them in time. Iteration itself remains valuable, since every strategy in the table replans and the poorest configurations are those whose limit forces short, repetitive plans; what fades is the specific benefit of changing the objective during the flight.

Objective scheduling is worth its complexity when the fleet is small, which is also the case where a wrong choice is most expensive, since a single aircraft has no second chance. As the fleet grows, a well-chosen fixed bound does the job, and the effort is better put elsewhere.

Two cautions go with this table. Each cell is a single run, as in the original experiment, so the ranking charts the space of strategies and makes no claim of statistical significance; a small gap between neighboring rows is not an ordering. And the configurations that reached 64 replans were stopped by that ceiling, not by the horizon, so their scores are lower bounds on what they would have reached.

\subsection{Replanning Frequency Analysis}

This section asks how often the route should be rebuilt, and what each rebuild does to plan quality and to the detection achieved. Every cadence was measured under two budget regimes, because the two answer different questions. Under a fixed total budget the whole mission gets one allowance of solver time, shared out among the replans; this tells us whether iteration pays at equal computational effort. Under a full budget per replan every planning step gets the entire allowance; this is the regime closest to operations, because between two decisions the aircraft is already in the air and that time passes whether the solver uses it or not.

\clearpage
\begin{table}[!ht]
\centering
\caption{Impact of replanning cadence under the two budget regimes. The Replans column gives the mean number of replans actually triggered in each regime, which is a consequence of the cadence and not a parameter.}
\label{tab:replanning_freq}
\begin{tabular}{lcrr}
\toprule
\textbf{Cadence} & \textbf{Replans} & \textbf{Fixed total budget} & \textbf{Full budget per replan} \\
\midrule
Never (static)     & 1.0 / 1.0   & 53.11 & 53.54 \\
\textbf{Every $H/2$} & 2.0 / 7.1 & \textbf{55.43} & \textbf{60.52} \\
Every $H/4$        & 4.0 / 10.0  & 40.49 & 52.41 \\
Every trip         & 8.0 / 13.4  & 52.15 & 60.36 \\
Every 15 obs.      & 1.5 / 8.5   & 46.84 & 59.83 \\
Every 5 obs.       & 4.0 / 15.2  & 47.31 & 58.52 \\
Every observation  & 17.0 / 56.1 & 44.80 & 53.41 \\
\bottomrule
\end{tabular}
\end{table}

Under the fixed budget the best cadence beats the static baseline by 2.3 points, and five of the six iterative cadences fall below it. The reason is mechanical: the more often the planner rebuilds the route, the thinner each solver slice becomes, and what the new information adds does not cover what the plan loses in quality. 

Under the operational regime replanning is clearly worth its cost. Four cadences beat the static baseline by 5 to 7 points. The contrast between the two columns is the useful measurement here, because it separates the effect of information from the effect of effort. 

The benefit is not monotonic in either regime. Replanning at every observation triggers 56 rebuilds and falls back to static levels. The best setting is a moderate one, replanning at the half-horizon, which balances updated environmental knowledge against enough solver depth to exploit it. Taken together with the previous section, this explains why replanning showed no isolated gain there: that experiment ran with a smaller per-replan allowance, and the balance had not yet tipped. 

\section{Conclusion}
\label{sec:concl}

We presented a route-planning framework for search and rescue that turns a probability grid over a disaster area into flyable routes for a UAV fleet. The problem was formulated as a probabilistic model of the multi-depot OVRP problem, in two complementary versions, maximising collected probability under a travel limit, and minimising travel under a coverage floor, and solved with a RKO whose decoder maps key vectors into multi-visit routing plans. An exact formulation solved with a commercial solver provides the reference against which the heuristic is measured.

On small instances the exact method proves optimality and the RKO reaches those optima; as the instances grow the solver stops converging within the time limit while the heuristic keeps producing usable plans within the 120-second limit. This is the regime that matters operationally, where a prompt workable plan outweighs a provably best one.

Under a fixed total allowance, slicing time among rebuilds costs more than the new information is worth, and iteration is close to neutral, however could assist with operations defined by the specialist when setting the search parameters. Under the allowance a real mission actually offers, planning time that elapses while the aircraft flies, replanning gains 5 to 7 percentage points. The benefit is not monotonic in frequency: a moderate cadence is best, and rebuilding at every observation performs no better than never rebuilding at all.

With one aircraft, switching from probability maximisation to travel minimisation midway through the mission gains twelve points over the best fixed setting. With two aircraft the advantage vanishes, and with three the best plan of the entire grid is a fixed one. Extra aircraft appear to supply the same flexibility the schedule was supplying, since several routes in the air can pursue different aims simultaneously rather than alternating between them in time. The level of the constraint, by contrast, matters everywhere: within a single objective, moving only the limit carries the score across almost the whole range the instance allows. Calibration effort belongs on the limit before it goes on the schedule.

Noisy probability maps are absorbed; biased ones are not. Independent noise on every cell still leaves a map worth flying with, whereas a map that is wrong in a consistent direction does far worse than no map at all. Operationally, an unreliable heat map is better discarded than trusted.

To the next steps: The planner does not revise its belief about where the victims are, so a Bayesian update of the map after each empty observation is the natural extension. The reward is linear in the number of visits, which overstates repeated observations, and detection in a visited cell is assumed certain; both should be relaxed, together with the two-dimensional, obstacle-free corridors, before field trials. The strategy grid was flown once per cell and some configurations hit the replan ceiling, so that ranking needs replication. Letting the system choose objective and constraint from the state of the mission, and validating the framework on established benchmarks and in real operations, complete the agenda.

\backmatter

\section*{Declarations}

\begin{itemize}
\item \textbf{Funding:} This study was financed in part by the Coordenação de Aperfeiçoamento de Pessoal de Nível Superior - Brasil (CAPES) - Finance Code 001.
\item \textbf{Conflict of interest:} The authors declare no competing interests.
\item \textbf{Ethics approval and consent to participate:} Not applicable, as this study is a computational simulation, no human or animal subjects.
\item \textbf{Consent for publication:} Not applicable.
\item \textbf{Materials availability:} Not applicable.
\item \textbf{Data and Code availability:} The code and data are available at \url{https://github.com/AntonioDantas/saruav_gurobi_python}.
\item \textbf{Author contribution:} A.D. conceived the framework, implemented it and carried out the computational experiments. E.S. developed the integration with the UAV platform and led the application of the framework to the search-and-rescue setting. A.C. formulated the mathematical model and adapted the framework accordingly. A.D. wrote the first draft of the manuscript, and all authors commented on previous versions. All authors read and approved the final manuscript.
\end{itemize}

\bibliography{sn-bibliography}

\end{document}